\documentclass{article}
\usepackage{amsmath}
\usepackage{graphicx}
\usepackage{float}
\usepackage{booktabs}
\usepackage{adjustbox}
\usepackage[margin=1in]{geometry}
\begin{document}

\title{Language Models Are An Effective Representation Learning Technique For Electronic Health Record Data}

\author{
Ethan Steinberg, Ken Jung, Jason A. Fries, Conor K. Corbin, Stephen R. Pfohl, Nigam H. Shah
}

\maketitle

\begin{abstract}
Widespread adoption of electronic health records (EHRs) has fueled the development of using machine learning to build prediction models for various clinical outcomes. This process is often constrained by having a relatively small number of patient records for training the model. We demonstrate that using patient representation schemes inspired from techniques in natural language processing can increase the accuracy of clinical prediction models by transferring information learned from the entire patient population to the task of training a specific model, where only a subset of the population is relevant. Such patient representation schemes enable a 3.5\% mean improvement in AUROC on five prediction tasks compared to standard baselines, with the average improvement rising to 19\% when only a small number of patient records are available for training the clinical prediction model.
\end{abstract}

\section*{Keywords}
Electronic health record, representation learning, transfer learning, risk stratification, machine learning

\section{Introduction}

The widespread use of electronic health records (EHRs) combined with the power of machine learning has the potential to reduce healthcare costs and improve quality of care \cite{Shilo2020, Norgeot2019, Weins2019,2020}. EHR data has been used to learn prediction models for outcomes such as mortality \cite{avati}, sepsis \cite{sepsis}, future cost of care \cite{Tamange011580}, 30-day readmission \cite{readmission} and others \cite{Google, FH}. The outputs of these clinical prediction models facilitate risk stratification and targeted intervention to improve the quality of care. \cite{kaiser, sepsisDeploy}. To date, most clinical prediction models use a small number of features and are trained using a small number of patient records \cite{Goldstein2017}.

The complexity of EHRs poses many obstacles for training clinical prediction models. EHR data is variable length, high dimensional and sparse, with complex temporal and hierarchical structure. They are comprised of irregularly spaced visits spread across years, with each visit containing a subset of thousands of possible diagnosis, procedure, and medication codes, as well as laboratory test results, unstructured text, and images. In contrast, most off-the-shelf machine learning algorithms expect a fixed length vector of features as input. Defining a transformation of patient records into such a fixed length representation is often a manual process that is time consuming and task-dependent, leaving much of the temporal and hierarchical structure of EHRs underutilized when training clinical prediction models.

Recent work on training clinical prediction models has used deep neural networks in an attempt to leverage the information inherent in the structure of EHRs, to directly capture the structure of medical data while training the model for a given clinical outcome (e.g., mortality or 30 day readmissions) \cite{Google}. Such an ``end-to-end" formulation is appealing because it has led to ground-breaking accuracy in computer vision and natural language processing (NLP) without requiring manual feature engineering.  However, this approach does not seem to provide consistent gains when applied to electronic health records. Comparisons with simple yet strong baselines \cite{Google, ChenNPJ2019} found that end-to-end neural network models provide minimal or no accuracy advantage over count-based representations combined with logistic regression or gradient boosted trees. One possible explanation for this limited improvement is that deep learning models typically require large training datasets and EHR datasets are limited by the number of patients with a given outcome in a particular health system's data.

Researchers in NLP and computer vision, when faced with small datasets, often use the technique of transfer learning to achieve gains in accuracy in small data situations \cite{umlfit, bert}. Transfer learning posits that it is possible to train a model for one task on a large dataset and then fine tune that model for a different task using a smaller dataset in order to achieve better performance on the second task than would be achieved by training a model \textit{de novo}. The choice of task used for pre-training is critical, with much of the current work in transfer learning focusing on designing a good task that helps capture useful structure that can be shared across tasks \cite{bert}. One common pre-training task that performs reasonably well for natural language processing is language modeling. Language modeling consists of learning a generative sequence model for text. After performing pre-training, the learned information stored in that model then needs to be transferred to a particular task of interest. \textit{Representation learning} is a type of transfer learning that focuses on performing that transfer by constructing fixed length representations which are then reused for downstream tasks. Transfer learning is especially compelling for training clinical prediction models using EHR data because it is often the case that the number of patients available in a training set for a given outcome is a small fraction of all the patients in an institution's EHR system \cite{Wiens2014}.

Our core hypothesis is that it is possible to use data from large numbers of patients to learn reusable, fixed length representations that improve the accuracy of clinical prediction models trained on smaller subsets of patients. There has been some prior work on applying representation learning methods to EHR data \cite{deeppatient, med2vec, otherw2v, edwardw2v}. However, these proposed representation learning techniques only capture parts of the EHR (such as visits \cite{med2vec} or codes \cite{otherw2v, edwardw2v}), and the relative performance of these methods against each other and against simple, count based representations is unknown. 

\begin{figure}[!b]
\centerline{
\includegraphics{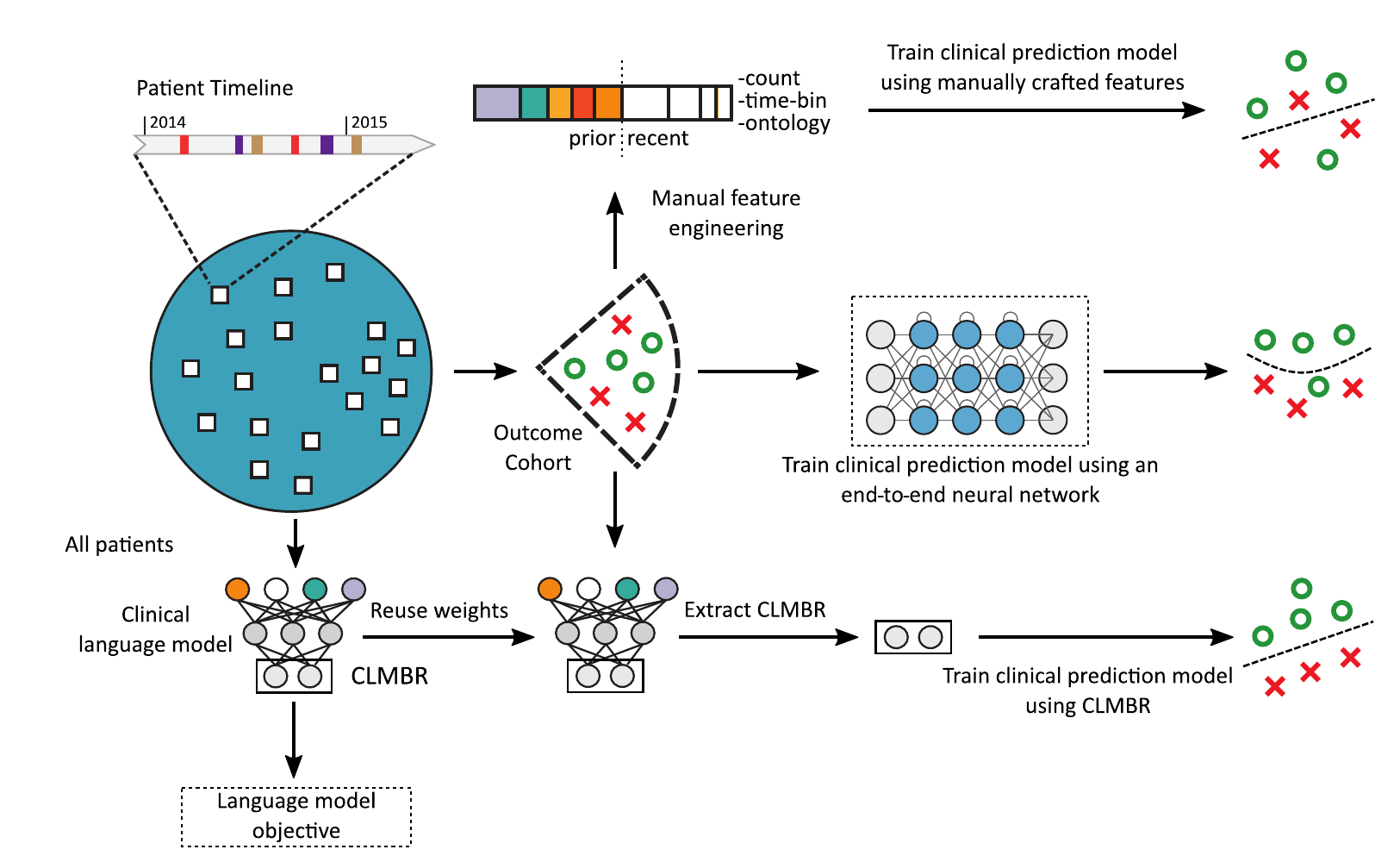}
}
\caption{An overview of the different approaches to training a predictive model for clinical outcomes: feature engineering, end-to-end neural network modeling, and representation learning through an approach such as clinical language modeling based representations (CLMBR).}
\label{representation}
\end{figure}

In this work we propose an improved generative sequence model for EHR data (a ``clinical language model") and show that this clinical language model can be used to derive representations in an approach we refer to as \textit{clinical language model based representations} (CLMBR). We empirically evaluated the effectiveness of this approach for training models on five prediction tasks as compared to published representation learning techniques. We compared clinical prediction models trained using CLMBR with clinical prediction models trained using simple count representations and with end-to-end trained deep neural networks for the same outcomes, as illustrated in Figure \ref{representation}. We investigated how the performance gains on clinical prediction models using learned representations varied as a function of the amount of data available for training the clinical prediction model. Finally, we showed how the clinical language model used in this work provides better representations than a previously published clinical language model \cite{doctorai}. 

\subsection{Related Work}
\subsubsection{Deep Neural Network Based Clinical Prediction Models Using EHR Data}
Recent work on training clinical prediction models using EHR data focuses on deep neural network models trained in an end-to-end manner for an outcome of interest. Clinical prediction models have been built for many outcomes, such as all-cause mortality \cite{avati}, heart failure \cite{GRAM, ChoiJAMIA2016HF, retain}, COPD \cite{Cheng2016RiskPW}, unplanned readmissions \cite{pham2016deepcare, nguyen2016mathtt}, and future hospital admissions \cite{zhang2018patient2vec}. These efforts generally propose novel neural net architectures and report performance gains over baseline models.  However, Rajkomar et al \cite{Google} reported that logistic regression using a simple bag of words based representation performed very close to or within the margin of error of an ensemble of three complex neural net models for three outcomes (inpatient mortality, readmissions and long length of stay). Similarly, Chen et al \cite{ChenNPJ2019} found that neural networks were consistently outperformed by gradient boosted trees and random forests on a range of clinical outcomes.  These seemingly conflicting results require further investigation of the benefits of using neural networks.  

\subsubsection{Representation Learning}
Representation learning is used in computer vision and NLP to mitigate the impact of limited training data \cite{umlfit}. Prior work on representation learning for EHR data primarily follows work in natural language processing because of similarities in the structure of data. For example, a document in natural language can be viewed as a sequence of words, and representations can be learned for either single words or entire sequences. Analogously, a patient's longitudinal EHR can be seen as a ``document" consisting of a sequence of diagnosis, procedure, medication, and laboratory codes. Note that this discussion is not about processing the textual content of clinical notes via natural language processing.

Representation learning for documents in natural language settings commonly focuses on learning word and document level representations. \emph{Word level representations} are fixed length vectors for each word learned through information theory and linear algebra \cite{pennington2014glove} or neural networks \cite{word2vec}. Here, the aim is to learn a representation that anticipates surrounding context words (e.g., in this sentence, the context of ``representation" includes ``learn" and ``anticipates"). The end result is a fixed length vector representation of each word which can then be used for tasks such as question answering and sentiment analysis \cite{swem, textcnn}. In contrast, \emph{document level representations} are fixed length vectors that capture salient properties of the whole document. A classic technique for doing so is Latent Semantic Indexing (LSI) \cite{lsi}, which combines both singular value decomposition and term frequency-inverse document frequency to learn a low dimensional vector representation of a document with the goal of maximizing the ability to reconstruct document term frequencies.

Currently, the most effective representation learning techniques for natural language focus on building better document level representations by learning language models. A \emph{language model} is a probabilistic model of sequences of words, often formulated as a neural network with millions of parameters that capture (or \emph{model}; hence the phrase language model) the language generation process by predicting a word at a time, either sequentially with recurrent neural networks \cite{umlfit} or via masking with transformer models \cite{bert}.

\vspace{5mm} 
\noindent \emph{Representation Learning For Electronic Health Records} \\
Analogous to word level representations, it is possible to treat medical codes in the EHR as words and learn representations for medical codes by adapting word2vec to deal with the lack of ordering of medical codes within an encounter \cite{otherw2v, edwardw2v}. Choi et al \cite{edwardw2v} used the code vectors to learn models that predict heart failure.  Extending to document level representations, in follow up work, Choi et al simultaneously learned medical code and patient level representations \cite{med2vec}. However, later evaluations found this approach was only a little better than several other baselines in predicting congestive heart failure \cite{MiME}. Miotto et al \cite{deeppatient} learned patient level representations using autoencoders, reporting significantly better performance for training models that predict future diagnosis codes over the next year.  However, in Choi et al \cite{med2vec}, stacked autoencoders were found to be no better than other baselines at predicting the next encounter's diagnosis codes.  

Researchers have also applied language modeling to EHRs. Prior work by Choi et al \cite{doctorai} proposed a language model (named DoctorAI) that predicts a subset of medical codes appearing in a sequence of patient encounters. They reported that a simple Gated Recurrent Unit (GRU) architecture performs quite well for this task. The DoctorAI language model used high level (i.e., 3 digit) diagnosis and medication codes and did not use laboratory tests or procedure codes. These choices enabled DoctorAI to make assumptions allowing a softmax probability transformation and a flat code output space to reduce computational complexity. They measure how well this language model captures the series of codes in EHR data and show that a GRU does better than several simpler baselines. However, Choi et al never evaluated whether such a language model could be used to improve performance on clinical prediction models.  Therefore, the utility of learning general purpose representations of EHR data for developing more accurate clinical prediction models remains unclear.

\section{Materials and Methods}
\label{methods}
\begin{figure}[ht]
\centerline{
\includegraphics{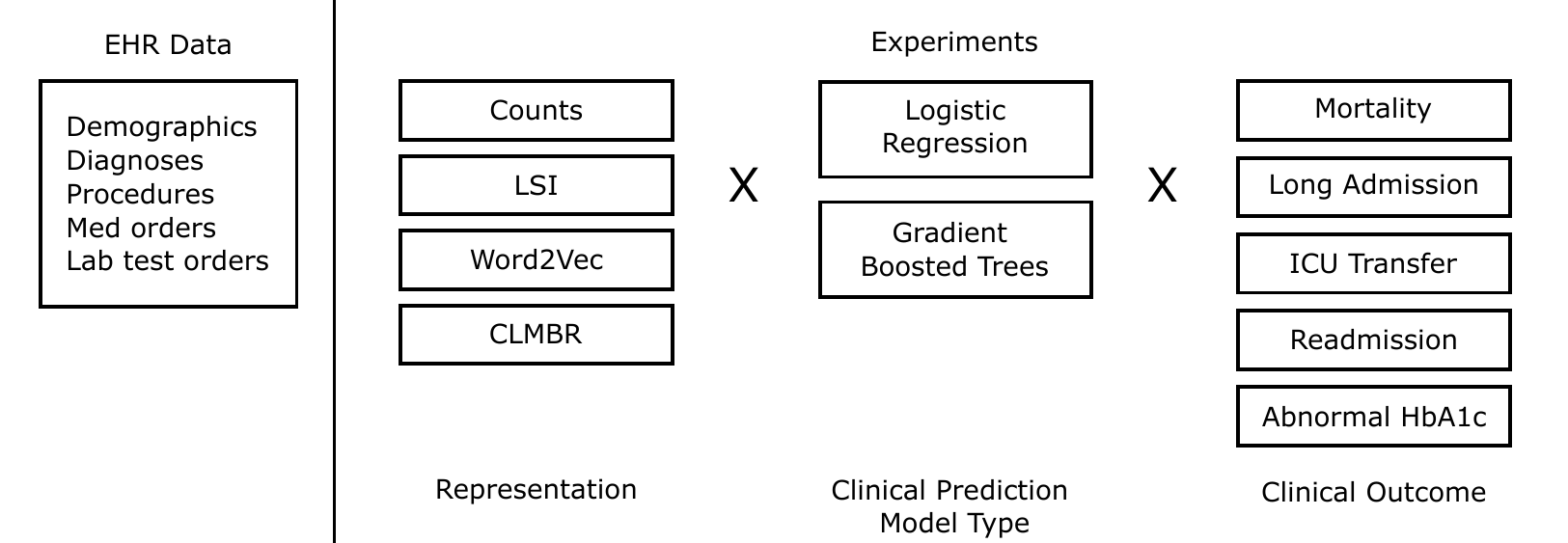}
}
\caption{Overview of experiments evaluating representation learning methods using EHR data. We evaluated four representation learning methods with two model types to train clinical prediction models for five outcomes.}
\label{benchmark}
\end{figure}

We evaluated the performance of four categories of representations (\emph{Counts}, \emph{Word2Vec}, \emph{LSI}, and \emph{CLMBR}) used as inputs to a logistic regression and to gradient boosted trees for predicting five outcomes.  Logistic regression and gradient boosted trees were chosen because they are widely used to train clinical prediction models and often perform quite well \cite{ChenNPJ2019}. As an additional baseline, we also report results of a clinical prediction model trained as an end-to-end GRU, which directly used the raw EHR data and internally learned a representation during the process of training for a particular clinical outcome. Figure \ref{benchmark} shows an overview of the experimental set up.

\subsection{Data}
All experiments were conducted on de-identified EHR data from Stanford Hospital and Lucile Packard Children's Hospital. The data comprises 3.4 million patient records spanning from 1990 through 2018.  The study was done with approval by Stanford University's Institutional Review Board.  We treated each patient's record as a sequence of days $d_1, \dots, d_N$, ordered by time.  Each day was associated with a set of medical codes for diagnoses, procedures, medication orders, and laboratory test orders (ICD10, CPT or HCPCS, RXCUI, and LOINC codes respectively) recorded on that day. In this study, we did not use quantitative information such as laboratory test results or vital sign measurements. We also did not use clinical notes, images, or explicit linkages between codes (e.g., diagnosis codes entered to justify procedures, as used in Choi et al \cite{MiME}).  In total, there were 21,664 codes after filtering for codes that occurred in the records of at least 25 patients.  Patient demographic data (gender, race and ethnicity) was encoded by assigning corresponding codes to the date of birth of the patient. 

\subsection{Experimental Setup}
We compared the effectiveness of simple count based and learned representations in terms of the discrimination accuracy of predictive models for five clinical outcomes across a range of training set sizes. Table \ref{final_task_def} provides a description of the clinical outcomes; Table \ref{final_task} describes the dataset for each clinical outcome. 

\begin{table}[]
\centering
\caption{Definitions of Clinical Outcomes}
\label{final_task_def}
\begin{adjustbox}{max width=\textwidth}
\begin{tabular}{@{}l p{0.5\linewidth}p{0.3\linewidth}@{}}
\toprule
Outcome & Definition & Time of Prediction \\
\midrule
Inpatient Mortality  & A patient death occurring during an inpatient stay & At admission \\
Long Admission & A patient stay of seven or more days in the hospital & At admission \\
ICU Transfer & Transfer of the patient to the ICU the following day & Every day of an inpatient stay \\
30-day Readmission & A patient readmitted to the hospital within 30 days & At discharge \\
Abnormal HbA1c & An HbA1c value $> 6.5\%$ for a non-diabetic patient &  Before the test result is returned \\ 
\bottomrule
\end{tabular}
\end{adjustbox}
\end{table}

\begin{table}[tb]
\centering
\caption{Characteristics of the dataset for each clinical outcome}
\label{final_task}
\begin{tabular}{@{}lrrr@{}}
\toprule
Outcome Name & Num Labels & Num Positives & Num Unique Patients \\
\midrule
Inpatient Mortality & 212,599 & 4,294  & 130,708 \\
Long Admission & 212,636 & 48,508 & 130,719 \\
ICU Transfer & 761,658 & 8,094 & 101,999 \\
30-day Readmission & 187,866 & 29,693 & 112,264  \\
Abnormal HbA1c & 83,550 & 1,651 & 51,654 \\
\bottomrule
\end{tabular}
\end{table}

\subsubsection{Data Splits}
Data was split into training, development, and test sets by time: data through December 31, 2015 was used for training, data from January 1, 2016 through July 1, 2016 was used for hyperparameter tuning, and data from August 1, 2016 through August 1, 2017 was used as a held out test set.  We adopted this design because of potential non-stationarity in EHRs, such that this scheme provides a more unbiased estimate of real world performance than time-agnostic patient splits  \cite{WiensCautionaryTale}.  Note that even though patients may have been included in multiple splits, examples consist of both a patient and a time of prediction, and the times of prediction do not overlap between the splits.  

\subsubsection{Clinical Prediction Models}
We used each representation (details in section 2.3) as input to two types of models to train a clinical prediction model for each clinical outcome.  The first was a simple linear model: logistic regression with $L_2$ regularization.  The $L_2$ strength parameter was swept in a grid for every power of 10 between $10^{-6}$ to $10^6$. We used the Sci-kit Learn's logistic regression implementation with the LBFGS algorithm \cite{scikit-learn}.  The second model type was gradient boosted trees, which can model interactions and non-linearities in the data.  We performed hyperparameter tuning by grid search, varying the learning rate between 0.02, 0.1, and 0.5; and the number of leaf nodes in each base tree between 10, 25, and 100.  Early stopping with 500 max trees was used for selecting the number of trees. We used the LightGBM \cite{lightgbm} implementation of gradient boosting.  

\subsubsection{Subsampling Experiments}
We also evaluated clinical prediction models trained using representations derived from smaller datasets to test the hypothesis that representation learning provides greater benefits as sample sizes decrease.  We performed experiments in which training and development sets were subsampled without replacement, with stratified sampling of the training and development sets to enforce a fixed positive label prevalence of 10\%.  The total sample sizes were 100, 200, 400, 800, 1,600, and 3,200, with 70\% and 30\% of each sample drawn from the training and development splits respectively.  The subsampled training and development splits were then used for clinical prediction model tuning and fitting. This process was repeated 10 times in order to provide estimates of variance due to sampling of the training and development sets for the performance metrics.

\subsubsection{End-to-End Neural Network Clinical Prediction Models}
To confirm the utility of general purpose learned representations, it is necessary to quantify the degree to which they differ from the end-to-end setup, especially for large sample sizes where end-to-end models tend to perform especially well.  To this end, we also trained end-to-end recurrent neural net models for each outcome. These models did not use any of the learned representations, and instead operated directly on the raw data itself, i.e., a sequence of observed codes. We used the same architecture as the language models except that the output of the GRU was fed directly into a simple logistic regression layer to predict the clinical outcomes. A hyperparameter search was performed independently for each outcome in order to provide a fair comparison. See Appendix \ref{end_to_end_grid} for the hyperparameter grid of the end-to-end GRU models. See Appendix \ref{Best_gru} for the best performing hyperparameters for each clinical prediction model. The difference with the general purpose representation was that each clinical prediction model could have learned a \textit{different} patient representation scheme.

\subsection{Representations}
\label{representations}
We examine four categories of representations for their utility in terms of the accuracy of predictive models for five clinical outcomes.

\subsubsection{Count Based Representations} \label{count}
The simplest representation we considered was counts of each code in the EHRs.  This representation is widely used as a baseline, and in Rajkomar et al \cite{Google} it resulted in excellent accuracy of regularized logistic regression based predictive models for three clinical outcomes.  

We also evaluated two enhancements to the basic counts representation: time binning and ontology expansion.  \emph{Time binning} counts occurrences of a code in different time buckets  separately, and has been used in prior work \cite{avati, Google}.  We used time buckets of 0-30 days, 30-180 days, 180-365 days, and 365+ days from the reference time.  These representations were very high dimensional and sparse because there are many codes, most of which occurred in a very few patients.  \emph{Ontology expansion} is a commonly used technique that mitigates this problem by using ontologies (knowledge bases that specify hierarchical relationships between concepts, e.g., ``Type 1 diabetes mellitus with ketoacidosis" is a type of ``Type 1 diabetes mellitus") to ``densify" these representations \cite{GRAM}.  For example, if we observed the ICD10 code E10.1 (``Type 1 diabetes mellitus"), we also counted that as an occurrence of the ancestor codes E10 (``Type 1 diabetes mellitus") and E08-E13 (``Diabetes mellitus").  We used the Unified Medical Language System (UMLS) \cite{umls} and mapped codes to their ancestors within their respective hierarchies when applicable (ICD10 for diagnoses, CPT or MTHH for procedures, and ATC for medications).  Note that this procedure increased the dimensionality of the representation to 36,617 codes because many ancestor codes were not present in the original representation.  

We thus evaluated four variations of count based representations, one for each combination of ontology expansion and time binning.  These representations were used as the baseline, comprising simple, non-learned representations.  

\subsubsection{Word2Vec Representation}
Adapting Word2Vec to patient EHR data requires managing the unordered nature of codes occurring on a given day.  Prior work \cite{edwardw2v, otherw2v} recommends randomly ordering the codes assigned on a given day into a sequence for input to word2vec. We implemented this strategy to construct embeddings for every code in our data, with an embedding size of 300 using gensim's word2vec implementation \cite{gensim}.  We also evaluated code embeddings generated from data augmented by ontology expansion as described above.  Finally, in order to construct patient level representations from the code embeddings, we evaluated combining code embeddings by taking the element-wise mean, and by concatenating the element-wise min, max and mean vectors as described in \cite{swem}.  These two ways of combining code emebeddings along with (or without) the ontology expansion resulted in four variations of word2vec based representations. 

\subsubsection{Latent Semantic Indexing Representations}
We applied LSI to construct patient level representations by treating each patient's EHR up to a randomly sampled time point as a ``document" in which each code is a ``word."  Following our count based representations \ref{count}, we ran LSI with and without ontology expansion and evaluated representation sizes of 400 and 800.  We thus evaluated four different LSI representations, again using gensim's implementation of LSI \cite{gensim}.

\subsubsection{Clinical Language Model Based Representations - CLMBR}
The core idea behind language model based representations is that they can capture global information about the sequence of tokens (such as words or disease codes) that is later useful for training a model to predict a clinical outcome. In EHRs, the sequences that we are trying to capture consist of days when a patient interacted with the health system. Mathematically, the language modeling objective is to estimate the probability of seeing a particular patient record, i.e., $p(d_1, \dots, d_N)$. To train the language model, probability distributions over sequences are factorized into a sequence of predictions where only a single element of the sequence is predicted at a time. Using EHR data, this corresponds to predicting the next day in a patient record given the previous days, i.e., $p(d_i | d_1, ... d_{i-1})$. Each $d_i$ is composed of a set of codes as opposed to a single token.  Thus, the problem is a multi-label prediction problem.  There is a large body of literature on multi-label problems \cite{}. For our experiments, we chose the simplest possible technique of transforming this problem into a binary classification problem through the binary relevance method \cite{BinaryRelevance}, modeled as

\[
p(d_i | d_1, ... d_{i-1}) = \prod_{c \in C} I(c \in d_i) p(c | d_1, ... d_{i-1}) + I(c \not \in d_i) (1 -  p(c | d_1, ... d_{i-1})).
\]
One issue with this approach is that this factorization will only be able to correctly model $p(d_i | d_1, \dots, d_{i-1})$ when the probability of each code is independent conditioned on the history $d_1, \dots, d_{i-1}$. That assumption is likely to be violated in EHR data, where there are strong correlations among certain codes that co-occur within an encounter. Nonetheless, we found that this approach seems to work well in practice.  

\begin{figure}[ht]
\centerline{
\includegraphics[width=0.9\textwidth]{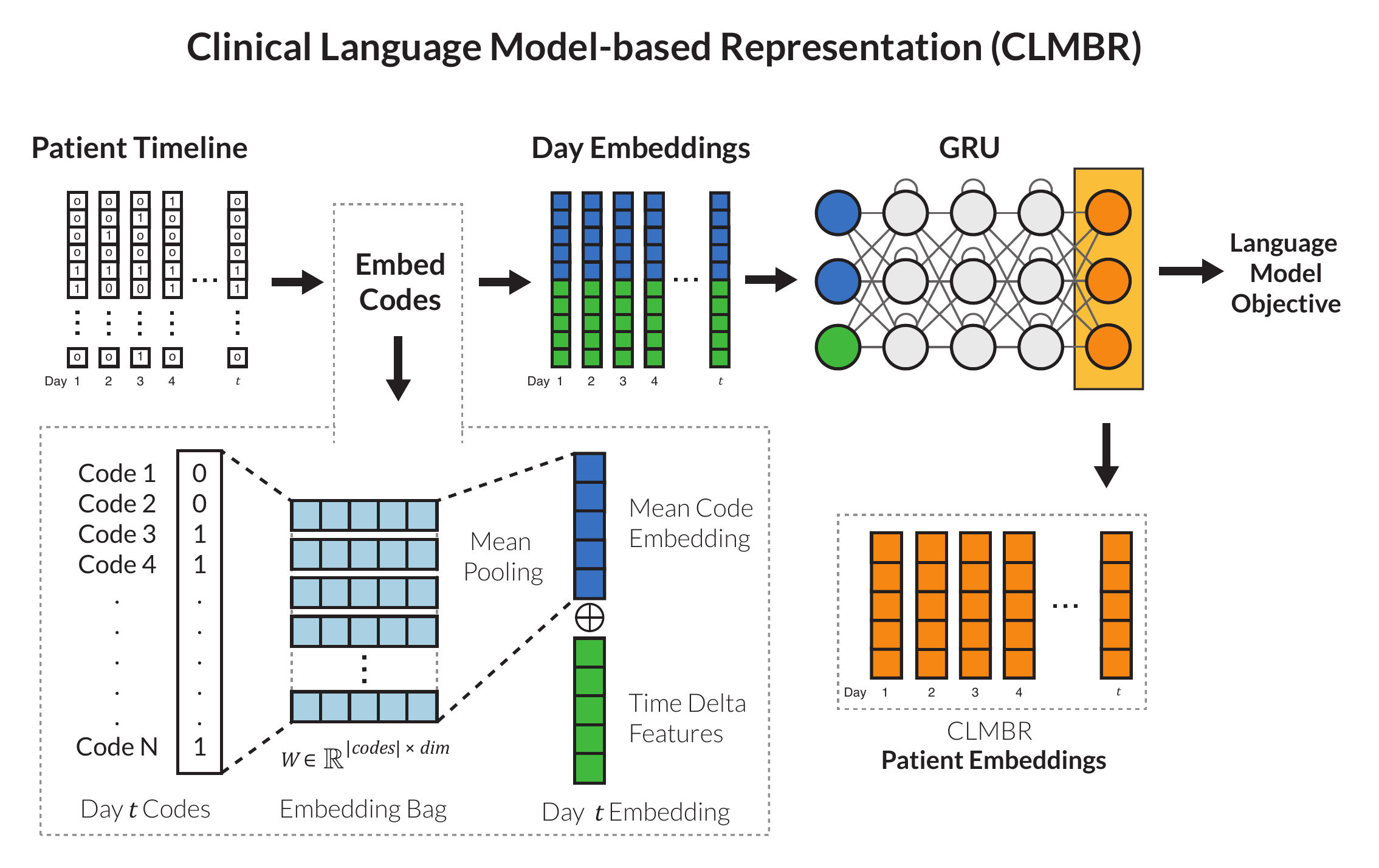}
}
\caption{The figure shows how patient representations were constructed using the CLMBR language model. Representations for individual patients were created by extracting fixed length vectors generated by the linear layer after the GRU.}
\label{language_model}
\end{figure}

Like prior work, we used a GRU-based neural network as our language model \cite{doctorai}. Figure \ref{language_model} shows an overview of the model architecture. The main modification we made is that we introduced a linear layer after the GRU layer in order to extract patient representations of a lower dimension from the internal GRU state. The first layer of our network was an embedding bag layer which took as input the sets of codes for each day and output the mean representation for that day using an embedding matrix $W$ with a tuned embedding size. As in Rajkomar et al \cite{Google}, day representations were then concatenated with a five element vector for each day that contained the age at a particular day, the log transform of the age at that particular day, the time delta from the previous day, the log transform of that time delta and a binary indicator of whether or not that day was the first day of the sequence. All variables were normalized to a mean of zero and a standard deviation of one. The purpose of adding these variables was to provide some time information to the neural network due to the fact that there were different amounts of real time passing between each day. The day representations plus demographic data were then fed into a single layer GRU with a set number of hidden units. A patient representation at each time step was computed by passing the output of the GRU through a GELU \cite{GELU} activation function and a linear layer with output size equal to the embedding size.

This patient representation was then used to compute the probability of each code in the code set in order to satisfy the language modeling objective. While it is ideal to compute these probabilities as a sigmoid transformation of the dot product between the patient representation and the code representation,  naively performing this computation is problematic due to both memory and computational needs caused by the large number of codes. Prior work \cite{hierarchy} has shown that it is possible to use a hierarchical decomposition for computing large scale softmax operations when there is pre-existing hierarchical structure. We applied a hierarchical decomposition to our code probability space using the ontologies in UMLS. We then applied that algorithm on a code matrix and a patient representation in order to obtain the probability of each code.  Following prior work on text language models, we used the same embedding matrix for both this computation and when computing our mean embeddings on the input side.  After the language model was trained, we extracted patient representations by taking the output of the linear layer prior to the hierarchical sigmoid layer.  More sophisticated approaches involving layer-wise fine-tuning were investigated, but did not appear to perform better than this simpler approach. Thus, we use the trained language model as a fixed feature extractor.  

We implemented this model in PyTorch and optimized it using OpenAI's version of the Adam algorithm \cite{Radford2018ImprovingLU} using $L_2$ regularization. We applied dropout between the input embedding and the GRU and between the GRU and the linear layer. Two models were evaluated: a small model with an embedding size of 400, and a larger model with an embedding size of 800. For each model, the learning rate, dropout rate, $L_2$ regularization strength, and hidden layer size were tuned using grid search. We trained each model for 50 epochs with linear learning rate decay to zero with a two epoch linear learning rate warmup. A batch size of 2,000 days (using as many patients as possible in a greedy manner) was used. Xavier initialization was used for the code representations and the default PyTorch initialization was used for the other parameters. The full grid of evaluated hyperparameters is specified in Appendix \ref{lm_grid}. The optimal hyperparameters used in the following experiments can be found in Appendix \ref{Best_lm}.

We also implemented and evaluated the language modeling objective used in DoctorAI \cite{doctorai} to study the effect of the choice of the language modeling objective. Following DoctorAI, we ignored the multi-label nature of the problem and instead used a softmax loss function. In addition, we mirrored DoctorAI by simplifying our target space by only predicting high level diagnosis (3 token ICD10) and medication codes (leaf ATC) as opposed to the full code space considered in our main language model. As in DoctorAI, we used a simplified flat softmax, as the reduced code space renders techniques like hierarchical softmax unnecessary. In the experiments comparing the two language model variants, we used a fixed embedding size of 800.

\subsection{Tuning and Evaluation}
We carried out our experiments in three stages --- language model tuning, clinical prediction model tuning, and evaluation of the clinical prediction models on held out test data --- carefully designed to prevent leakage from the test set into the training and development datasets used to develop the language and clinical prediction models. Language models such as DoctorAI and CLMBR have many hyperparameters that must be tuned. We tuned them by fitting language models with different hyperparameter settings to the training set and selecting optimal settings based on the language model loss on the development set.  The final language models were re-trained on the combined training and development set data using these optimal settings. The hyperparameters for the clinical prediction models were tuned in a corresponding manner through training on the training set and selecting the optimal settings based on AUROC on the development set. We performed a separate hyperparameter search for each clinical prediction model in order to fairly measure performance.  We then re-trained the clinical prediction models using the optimal settings on the combined training and development set data.  Finally, the re-trained clinical prediction models were evaluated on the held out test set.  We calculated uncertainty estimates of the resulting performance by taking 1,001 bootstrap samples of the test set. 

%

\section{Results}
Our work primarily aims to measure the extent that clinical prediction models that leverage language model based representations outperform those that rely on engineered features or simpler representation learning techniques. In addition, we explore whether the magnitude of this effect changes as a function of both the amount of labeled training data available and the class of supervised learning algorithm used for the clinical prediction model. We also explore the importance of our more complicated clinical language model compared to prior clinical language modeling work.

\subsection{Difference in Performance of Clinical Prediction Models with Different Representations}
We first evaluated the difference in performance of clinical prediction models trained using alternative representations when a large amount of training data was available. Each outcome in the pool of five clinical outcomes was chosen on the basis of having a large number of labels to both aid this analysis and to reduce the variance of our performance estimates.  Table \ref{large_scale_performance} shows the AUROC on the test set for each representation category when trained with all of the data, with the best performing representation presented in bold font. All performance metrics were calculated pair-wise, relative to the counts representation, in order to reduce variance and better quantify differences between representations.  We report standard deviations estimated by 1,001 bootstrap samples of the test set.  Appendix \ref{Best_task} lists the best hyperparameter settings for each outcome and representation combination.  We found that models trained using CLMBR representations performed best for all five outcomes, although the improvement over alternatives was minimal for some of the outcomes.  Surprisingly, models trained using CLMBR representations were uniformly superior to the end-to-end GRU models.  Word2vec and LSI representations on the other hand were usually worse than other representations.

\begin{table}[!b]
\caption{Difference in AUROC of clinical prediction models trained on different representations}
\centering
\begin{adjustbox}{max width=\textwidth}
\begin{tabular}{@{}lrrrrr@{}} 
\toprule
    & {} & \multicolumn{4} {c} {Relative Compared To Counts Baseline} \\
 \cmidrule(lr){3-6}
Outcome Name & Counts & Word2Vec & LSI & CLMBR & End-to-end GRU \\ \midrule
Inpatient Mortality & $0.834$ & $-0.010 \pm 0.006 $ & $-0.046 \pm 0.007$ & $\boldsymbol{0.018 \pm 0.006}$ & $-0.030 \pm 0.008$ \\
Long Admission & $0.783$ & $-0.020 \pm 0.002$ & $-0.055 \pm 0.002$ & $\boldsymbol{0.009 \pm 0.002}$ & $-0.013 \pm 0.002$ \\
ICU Transfer & $0.792$ & $-0.041 \pm 0.006$ & $-0.086 \pm 0.007$ & $\boldsymbol{0.045 \pm 0.005}$ & $\hphantom{-}0.039 \pm 0.006 $ \\
30-day Readmission & $0.809$ & $-0.018 \pm 0.002$ & $-0.051 \pm 0.003$ & $\boldsymbol{0.005 \pm 0.002}$ & $-0.001 \pm 0.002  $ \\
Abnormal HbA1c & $0.700$ & $\hphantom{-}0.015 \pm 0.015 $ & $-0.011 \pm 0.016$ & $\boldsymbol{0.056 \pm 0.013}$ & $-0.019 \pm 0.017 $ \\
\bottomrule
\end{tabular}
\end{adjustbox}
\label{large_scale_performance}
\end{table}

\subsection{Effect of Training Set Size}
\begin{figure}[H]
\centerline{
\includegraphics{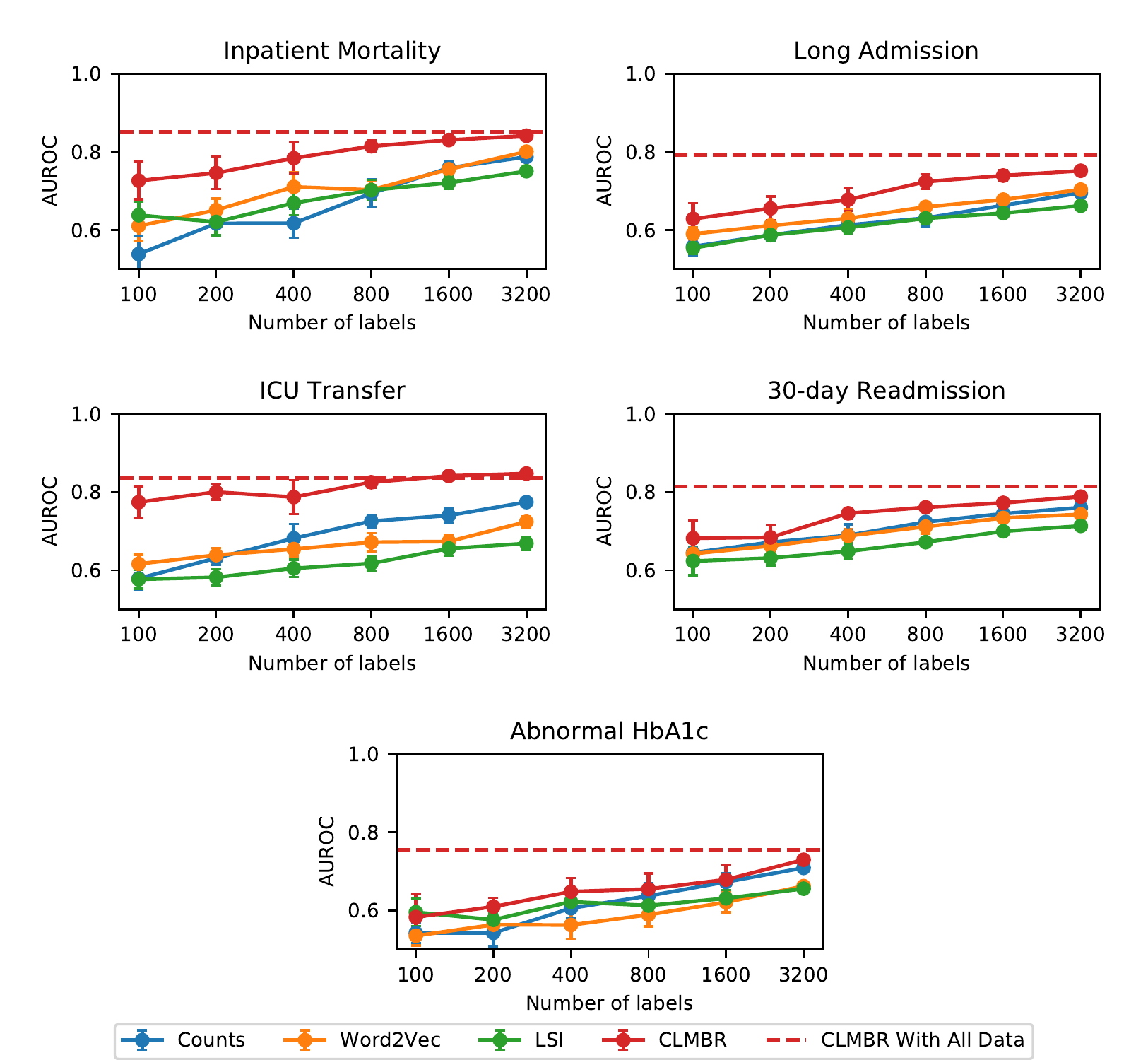}
}
\caption{AUROC (y-axis) for five clinical prediction models. For each clinical prediction model, the AUROC of models trained using a given representation type is plotted as a function of training set size (x-axis). Note that CLMBR (red) matched or outperformed all other approaches. In each plot, the dashed line shows performance of clinical prediction models trained using the CLMBR representation with the full dataset and represents best case performance.  We found that using the CLMBR representation increased the AUROC of the clinical prediction model for all outcomes and training set sizes, but the magnitude of the benefit was larger at smaller sample sizes and diminished at larger sample sizes.}
\label{subsampling}
\end{figure}

We also performed experiments in which we artificially reduced the dataset set sizes used for training clinical prediction models through subsampling. These experiments explored the hypothesis that learned representations are especially effective when there is only a limited amount of data available for training a clinical prediction model.  Figure \ref{subsampling} shows the AUROC of the clinical prediction models trained using different representations as dataset size changes. We calculated the mean AUROC across ten subsamples of the training set and show the 95\% t-distribution confidence interval for the mean.  Across all representation choices, we found that the AUROC decreased as the data set decreased in size. However, as expected, models trained using CLMBR representations fared best. The other learned representation classes, word2vec and LSI, seemed to provide some benefit relative to count based representations at smaller sample sizes (with word2vec outperforming LSI).  

\subsection{Effect of the Type of the Prediction Model as a Function of the Representation}
\begin{figure}[H]
\centerline{
\includegraphics{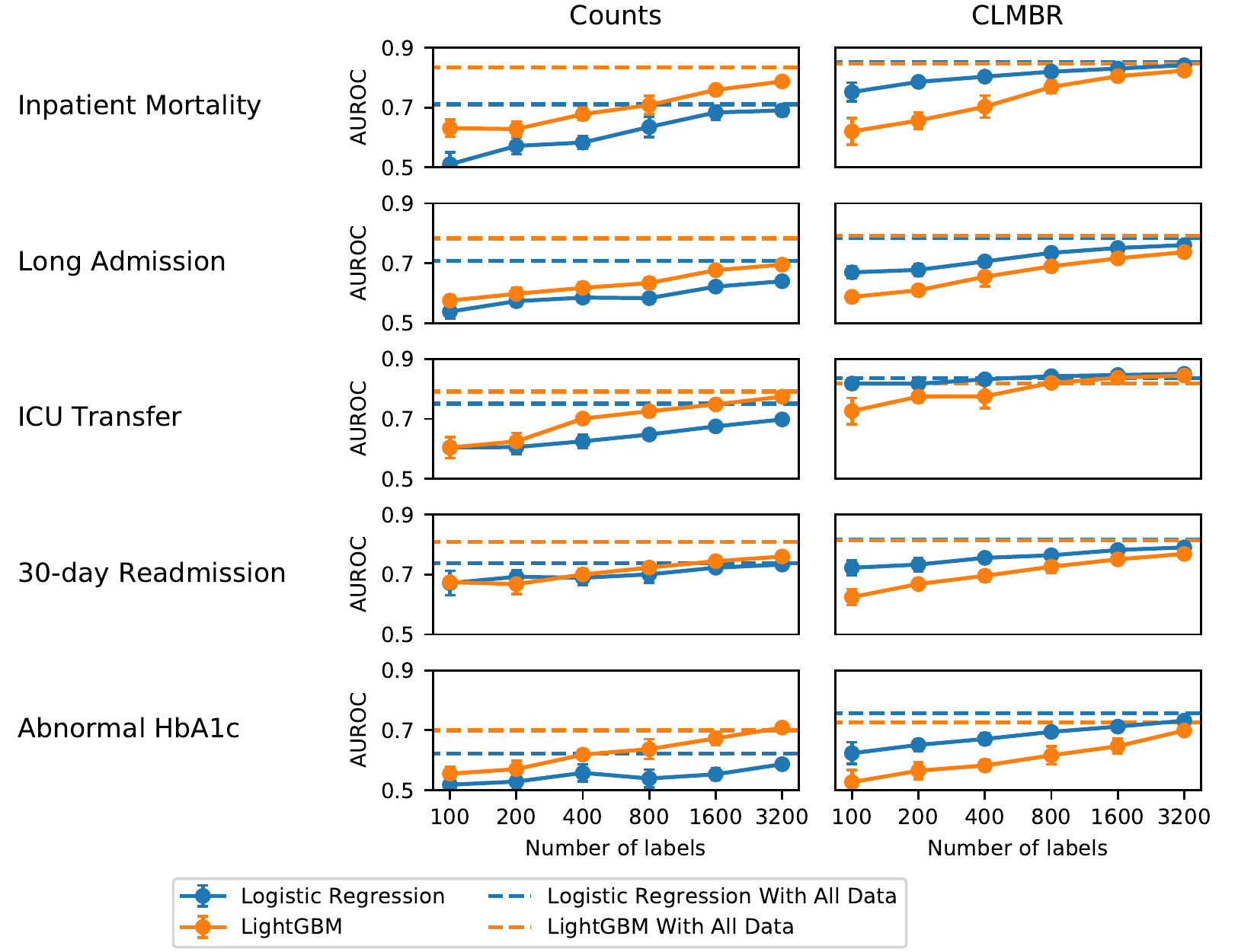}
}
\caption{AUROC for logistic regression and gradient boosted tree clinical prediction models trained using count based and CLMBR representations. For count based representations, we observed significant benefits from using gradient boosted trees versus $L_2$ regularized logistic regression. In contrast, we found that with CLMBR logistic regression outperformed gradient boosting models. The dashed lines show performance of the clinical prediction models trained on the full datasets using CLMBR representations and represents best case performance given available data.}

\label{logistic_lightgbm}
\end{figure}

We evaluated $L_2$ regularized logistic regression and gradient boosted tree for training the clinical prediction models, to identify which of these two performed best with which types of representations. This analysis is important because simpler models such as logistic regression are easier and faster to train than gradient boosted trees. Figure \ref{logistic_lightgbm} shows the relative performance of these model types for all five outcomes, over multiple sample sizes for count based representations and CLMBR.  As before, we computed the mean AUROC from ten subsamples of the training set and report the 95\% t confidence interval for that mean. When using count based representations, there was a consistent performance benefit in using gradient boosted tree models versus logistic regression, with gaps ranging from 4\% for 30 day readmission to 20.7\% for inpatient mortality.  With CLMBR representations, the best performing clinical prediction model type was a logistic regression model. More complex gradient boosted tree models offered no improvement even with large sample sizes, and often hurt performance at smaller sample sizes.

\subsection{Performance Difference Between CLMBR's Language Model and DoctorAI}
CLMBR's language model and DoctorAI \cite{doctorai} are both clinical language models. In order to measure the benefits of CLMBR's more complicated language model we implemented the DoctorAI language model as described in the methods and used it to construct representations. We then trained clinical prediction models using representations from the two different language models. Table \ref{simplified} shows the AUROCs of the prediction models along with the difference between the two and the standard deviation of that difference computed from bootstrap samples.  We observed a consistent improvement in performance across all outcomes with the complicated language modeling objective, with a larger improvement for two of the five outcomes (ICU transfer, abnormal HbA1c). 

\begin{table}[tb]
\caption{AUROC of prediction models with different language modeling objectives}
\centering
\begin{tabular}{@{}lrrr@{}} 
\toprule
Outcome Name & CLMBR Language Model & DoctorAI Language Model & Difference \\ \midrule
Inpatient Mortality & $0.852$ & $0.844$ & $-0.008 \pm 0.003$ \\
Long Admission & $0.792$ & $0.788$ & $-0.004 \pm 0.001$ \\
ICU Transfer & $0.837$ & $0.813$ & $-0.024 \pm 0.002$ \\
30-day Readmission & $0.814$ & $0.807$ & $-0.007 \pm 0.002$ \\
Abnormal HbA1c & $0.756$ & $0.742$ & $-0.014 \pm 0.008$ \\
\bottomrule
\end{tabular}
\label{simplified}
\end{table}

\section{Discussion}
Prior work employing deep neural networks to train prediction models for clinical outcomes using EHR data has focused mostly on end-to-end prediction models and used large datasets \cite{Google}. There has been much less work on learning general purpose representations using the entire EHR dataset that can then be re-used to train better prediction models. We have shown that language model based representations (such as DoctorAI and CLMBR), which capture the sequential nature of EHR data, are significantly better than a wide array of alternative representations for training clinical prediction models across a range of training set sizes. We also determined that the choice of the language model objective does matter, with the more expansive language model, CLMBR, providing better representations. The benefits of a language model based representation such as CLMBR are largest with small sample sizes (with an average improvement of 19\% in AUROC), but also hold with quite large sample sizes, including when training clinical prediction models with over 200,000 samples. 

Somewhat surprisingly, language model based representations also proved superior to end-to-end trained neural nets in the large sample regime. In contrast, other learned representations proved to be of little value relative to simpler count based representations when enough data was available. Finally, we found that with enough training data, clinical prediction models using simple count based representations can perform very well, being only 3.5\% worse than models trained using CLMBR representations.  However, with count based representations, it is important to use a model type with sufficient expressive power. Note that gradient boosted trees performed much better than $L_2$ regularized logistic regression with the simple counts based representations.  This observation may explain some of the discrepancies in reported performance gaps between deep neural network models and baselines in prior work \cite{Google, ChenNPJ2019, miotto2016deep, GRAM}. 

Currently, language model based representations come with significant upfront computation costs to train and tune \cite{bert}. However, this process is a one time cost per institution and can be amortized over many clinical outcomes. Moreover, recent language model work has demonstrated considerable reduction in training costs \cite{ELECTRA}.

Our conclusions have important limitations.  First, our findings are limited to the five clinical outcomes used in this work and findings may not generalize to all other possible EHR-based model types.  Second, this work does not explore how well CLMBR representations learned from data from one institution will generalize to other sites. In addition, we can expect the volume of EHR data available for training clinical prediction models to increase steadily, which might erase the gains from using more complex representation regimes. In particular, we note that end-to-end neural net models may regain the advantage when more training data is available.

\section{Conclusion}
In this work we developed and evaluated language model based representations for EHR data and found that the resulting patient representations were better than three other representation schemes as well as end-to-end neural network models for training prediction models for a variety of clinical outcomes at varying dataset sizes. The improvement in accuracy was especially significant at small sample sizes, with an average improvement of 19\% in AUROC at the smallest sample sizes. We also found that logistic regression models worked particularly well with language model based representations, potentially enabling faster and cheaper development of models for predicting clinical outcomes. These results suggest that language model based representations are a useful technique for developing better clinical prediction models using EHR data.

\section*{Acknowledgments}
This work was funded under NLM R01-LM011369-05. GPU resources were provided by Nero, a secure data science platform made possible by the Stanford School of Medicine Research Office and Stanford Research Computing Center. We would also like to thank Erin Craig, Agata Foryciarz and Sehj Kashyap for providing useful comments on the paper.

\section*{Author Information}
\subsection*{Affiliations} 

Stanford University, Stanford, CA, USA: \\
Ethan Steinberg, Ken Jung, Jason A. Fries, Conor K. Corbin, Stephen R. Pfohl, Nigam H. Shah

\subsection*{Contributions}

E.S. designed and conducted the primary experiments and created the initial drafts of the manuscript. K.J. helped work on further drafts and helped contribute code for some experiments. C.C. and S.P. contributed code for some of the experiments. J.F. and N.S. provided helpful discussion and helped design some of the figures. All authors contributed to revising the paper.

\subsection*{Corresponding Author}

Correspondence to Ethan Steinberg (ethanid@stanford.edu).

\section*{Competing Interests}
The authors have no conflicts of interest to declare.

\bibliographystyle{ws-procs11x85}
\bibliography{ws-pro-sample}

\begin{thebibliography}{10}

\bibitem{Shilo2020}
S.~Shilo, H.~Rossman and E.~Segal, Axes of a revolution: challenges and
  promises of big data in healthcare, {\em Nature Medicine} {\bf 26}, 29
  (2020).

\bibitem{Norgeot2019}
B.~Norgeot, B.~S. Glicksberg and A.~J. Butte, A call for deep-learning
  healthcare, {\em Nature Medicine} {\bf 25}, 14 (January 2019).

\bibitem{Weins2019}
J.~Wiens, S.~Saria, M.~Sendak, M.~Ghassemi, V.~X. Liu, F.~Doshi-Velez, K.~Jung,
  K.~Heller, D.~Kale, M.~Saeed, P.~N. Ossorio, S.~Thadaney-Israni and
  A.~Goldenberg, Do no harm: a roadmap for responsible machine learning for
  health care, {\em Nature Medicine} {\bf 25}, 1337 (August 2019).

\bibitem{2020}
S.~K. M. G. M. N. K. C. W.~R. Mark P.~Sendak, Joshua~D’Arcy and S.~Balu, A
  path for translation of machine learning products into healthcare delivery,
  {\em {EMJ} Innovations}  (January 2020).

\bibitem{avati}
A.~Avati, K.~Jung, S.~Harman, L.~Downing, A.~Ng and N.~H. Shah, Improving
  palliative care with deep learning, in {\em 2017 IEEE International
  Conference on Bioinformatics and Biomedicine (BIBM)\/}, Nov 2017.

\bibitem{sepsis}
M.~B. Dhudasia, S.~Mukhopadhyay and K.~M. Puopolo, Implementation of the sepsis
  risk calculator at an academic birth hospital, {\em Hospital Pediatrics} {\bf
  8}, 243  (2018).

\bibitem{Tamange011580}
S.~Tamang, A.~Milstein, H.~T. S{\o}rensen, L.~Pedersen, L.~Mackey, J.-R.
  Betterton, L.~Janson and N.~Shah, Predicting patient {\textquoteleft}cost
  blooms{\textquoteright} in denmark: a longitudinal population-based study,
  {\em BMJ Open} {\bf 7}  (2017).

\bibitem{readmission}
P.~Cronin, J.~Greenwald, G.~C.~Crevensten, H.~Chueh and A.~Zai, Development and
  implementation of a real-time 30-day readmission predictive model, {\em AMIA
  Annual Symposium proceedings / AMIA Symposium. AMIA Symposium} {\bf 2014},
  424 (11 2014).

\bibitem{Google}
A.~Rajkomar, E.~Oren, K.~Chen, A.~M. Dai, N.~Hajaj, M.~Hardt, P.~J. Liu,
  X.~Liu, J.~Marcus, M.~Sun, P.~Sundberg, H.~Yee, K.~Zhang, Y.~Zhang,
  G.~Flores, G.~E. Duggan, J.~Irvine, Q.~Le, K.~Litsch, A.~Mossin, J.~Tansuwan,
  D.~Wang, J.~Wexler, J.~Wilson, D.~Ludwig, S.~L. Volchenboum, K.~Chou,
  M.~Pearson, S.~Madabushi, N.~H. Shah, A.~J. Butte, M.~D. Howell, C.~Cui,
  G.~S. Corrado and J.~Dean, Scalable and accurate deep learning with
  electronic health records, {\em npj Digital Medicine} {\bf 1}, p.~18  (2018).

\bibitem{FH}
J.~M. Banda, A.~Sarraju, F.~Abbasi, J.~Parizo, M.~Pariani, H.~Ison, E.~Briskin,
  H.~Wand, S.~Dubois, K.~Jung, S.~A. Myers, D.~J. Rader, J.~B. Leader, M.~F.
  Murray, K.~D. Myers, K.~Wilemon, N.~H. Shah and J.~W. Knowles, Finding missed
  cases of familial hypercholesterolemia in health systems using machine
  learning, {\em npj Digital Medicine} {\bf 2}, p.~23  (2019).

\bibitem{kaiser}
S.~S. Paulson, B.~A. Dummett, J.~Green, E.~Scruth, V.~Reyes and G.~J. Escobar,
  What do we do after the pilot is done? implementation of a hospital early
  warning system at scale, {\em The Joint Commission Journal on Quality and
  Patient Safety}   (2020).

\bibitem{sepsisDeploy}
D.~W. Shimabukuro, C.~W. Barton, M.~D. Feldman, S.~J. Mataraso and R.~Das,
  Effect of a machine learning-based severe sepsis prediction algorithm on
  patient survival and hospital length of stay: a randomised clinical trial,
  {\em BMJ Open Respiratory Research} {\bf 4}  (2017).

\bibitem{Goldstein2017}
B.~A. Goldstein, A.~M. Navar, M.~J. Pencina and J.~P.~A. Ioannidis,
  Opportunities and challenges in developing risk prediction models with
  electronic health records data: a systematic review, {\em Journal of the
  American Medical Informatics Association} {\bf 24}, 198 (May 2016).

\bibitem{ChenNPJ2019}
D.~Chen, S.~Liu, P.~Kingsbury, S.~Sohn, C.~B. Sorlie, E.~B. Haberman, J.~M.
  Naessens, D.~W. Larson and H.~Liu, Deep learning and alternative learning
  strategies for retrospective real-world clinical data, {\em Nature Digital
  Medicine} {\bf 2}  (2019).

\bibitem{umlfit}
J.~Howard and S.~Ruder, Universal language model fine-tuning for text
  classification, in {\em Proceedings of the 56th Annual Meeting of the
  Association for Computational Linguistics (Volume 1: Long Papers)\/},
  (Association for Computational Linguistics, Melbourne, Australia, July 2018).

\bibitem{bert}
J.~Devlin, M.~Chang, K.~Lee and K.~Toutanova, Bert: Pre-training of deep
  bidirectional transformers for language understanding, in {\em NAACL-HLT\/},
  2018.

\bibitem{Wiens2014}
J.~Wiens, J.~Guttag and E.~Horvitz, A study in transfer learning: leveraging
  data from multiple hospitals to enhance hospital-specific predictions, {\em
  Journal of the American Medical Informatics Association} {\bf 21}, 699 (July
  2014).

\bibitem{deeppatient}
R.~Miotto, L.~Li, B.~A. Kidd and J.~T. Dudley, {{D}eep {P}atient: {A}n
  {U}nsupervised {R}epresentation to {P}redict the {F}uture of {P}atients from
  the {E}lectronic {H}ealth {R}ecords}, {\em Sci Rep} {\bf 6}, p. 26094 (05
  2016).

\bibitem{med2vec}
E.~Choi, M.~T. Bahadori, E.~Searles, C.~Coffey, M.~Thompson, J.~Bost, J.~T and
  J.~Sun, Multi-layer representation learning for medical concepts, in {\em
  Proceedings of the 22nd ACM SIGKDD International Conference on Knowledge
  Discovery and Data Mining\/}, KDD '16 (ACM, New York, NY, USA, 2016).

\bibitem{otherw2v}
Y.~Choi, C.~Y. Chiu and D.~Sontag, Learning low-dimensional representations of
  medical concepts, {\em AMIA Joint Summits on Translational Science
  proceedings. AMIA Joint Summits on Translational Science} {\bf 2016}, 41 (Jul
  2016), 27570647[pmid].

\bibitem{edwardw2v}
E.~Choi, A.~Schuetz, W.~F. Stewart and J.~Sun, Medical concept representation
  learning from electronic health records and its application on heart failure
  prediction, {\em CoRR} {\bf abs/1602.03686}  (2016).

\bibitem{doctorai}
E.~Choi, M.~T. Bahadori, A.~Schuetz, W.~F. Stewart and J.~Sun, Doctor ai:
  Predicting clinical events via recurrent neural networks, in {\em Proceedings
  of the 1st Machine Learning for Healthcare Conference\/},  eds.
  F.~Doshi-Velez, J.~Fackler, D.~Kale, B.~Wallace and J.~Wiens, Proceedings of
  Machine Learning Research, Vol.~56 (PMLR, Children's Hospital LA, Los
  Angeles, CA, USA, 18--19 Aug 2016).

\bibitem{GRAM}
E.~Choi, M.~T. Bahadori, L.~Song, W.~F. Stewart and J.~Sun, Gram: Graph-based
  attention model for healthcare representation learning, in {\em Proceedings
  of the 23rd ACM SIGKDD International Conference on Knowledge Discovery and
  Data Mining\/}, KDD '17 (ACM, New York, NY, USA, 2017).

\bibitem{ChoiJAMIA2016HF}
E.~Choi, A.~Schuetz, W.~F. Stewart and J.~Sun, Using recurrent neural network
  models for early detection of heart failure onset, {\em J Am Med Inform
  Assoc} {\bf 2}, 361  (2016).

\bibitem{retain}
E.~Choi, M.~T. Bahadori, A.~Schuetz, W.~F. Stewart and J.~Sun, {RETAIN:}
  interpretable predictive model in healthcare using reverse time attention
  mechanism, {\em CoRR} {\bf abs/1608.05745}  (2016).

\bibitem{Cheng2016RiskPW}
Y.~Cheng, F.~Wang, P.~Zhang and J.~Hu, Risk prediction with electronic health
  records: A deep learning approach, in {\em Proceedings of the 2016 SIAM
  International Conference on Data Mining\/}, 2016.

\bibitem{pham2016deepcare}
T.~Pham, T.~Tran, D.~Phung and S.~Venkatesh, Deepcare: A deep dynamic memory
  model for predictive medicine, in {\em Pacific-Asia Conference on Knowledge
  Discovery and Data Mining\/}, 2016.

\bibitem{nguyen2016mathtt}
P.~Nguyen, T.~Tran, N.~Wickramasinghe and S.~Venkatesh, Deepr: a convolutional
  net for medical records, {\em IEEE journal of biomedical and health
  informatics} {\bf 21}, 22  (2016).

\bibitem{zhang2018patient2vec}
J.~Zhang, K.~Kowsari, J.~H. Harrison, J.~M. Lobo and L.~E. Barnes, Patient2vec:
  A personalized interpretable deep representation of the longitudinal
  electronic health record, {\em IEEE Access} {\bf 6}, 65333  (2018).

\bibitem{pennington2014glove}
J.~Pennington, R.~Socher and C.~D. Manning, Glove: Global vectors for word
  representation, in {\em Empirical Methods in Natural Language Processing
  (EMNLP)\/}, 2014.

\bibitem{word2vec}
T.~Mikolov, I.~Sutskever, K.~Chen, G.~Corrado and J.~Dean, Distributed
  representations of words and phrases and their compositionality, in {\em
  Proceedings of the 26th International Conference on Neural Information
  Processing Systems - Volume 2\/}, NIPS'13 (Curran Associates Inc., USA,
  2013).

\bibitem{swem}
D.~Shen, G.~Wang, W.~Wang, M.~R. Min, Q.~Su, Y.~Zhang, C.~Li, R.~Henao and
  L.~Carin, Baseline needs more love: On simple word-embedding-based models and
  associated pooling mechanisms, in {\em Proceedings of the 56th Annual Meeting
  of the Association for Computational Linguistics (Volume 1: Long Papers)\/},
  (Association for Computational Linguistics, Melbourne, Australia, July 2018).

\bibitem{textcnn}
Y.~Kim, Convolutional neural networks for sentence classification, in {\em
  Proceedings of the 2014 Conference on Empirical Methods in Natural Language
  Processing ({EMNLP})\/},  (Association for Computational Linguistics, Doha,
  Qatar, October 2014).

\bibitem{lsi}
M.~W. Berry, S.~T. Dumais and G.~W. O'Brien, Using linear algebra for
  intelligent information retrieval, {\em SIAM Rev.} {\bf 37}, 573 (December
  1995).

\bibitem{MiME}
E.~Choi, C.~Xiao, W.~F. Stewart and J.~Sun, Mime: Multilevel medical embedding
  of electronic health records for predictive healthcare, {\em CoRR} {\bf
  abs/1810.09593}  (2018).

\bibitem{WiensCautionaryTale}
E.~Sherman, H.~Gurm, U.~Balis, S.~Owens and J.~Wiens, Leveraging clinical
  time-series data for prediction: a cautionary tale, in {\em AMIA Annual
  Symposium Proceedings\/},  (American Medical Informatics Association, 2017).

\bibitem{scikit-learn}
F.~Pedregosa, G.~Varoquaux, A.~Gramfort, V.~Michel, B.~Thirion, O.~Grisel,
  M.~Blondel, P.~Prettenhofer, R.~Weiss, V.~Dubourg, J.~Vanderplas, A.~Passos,
  D.~Cournapeau, M.~Brucher, M.~Perrot and E.~Duchesnay, Scikit-learn: Machine
  learning in {P}ython, {\em Journal of Machine Learning Research} {\bf 12},
  2825  (2011).

\bibitem{lightgbm}
G.~Ke, Q.~Meng, T.~Finley, T.~Wang, W.~Chen, W.~Ma, Q.~Ye and T.-Y. Liu,
  Lightgbm: A highly efficient gradient boosting decision tree, in {\em
  Proceedings of the 31st International Conference on Neural Information
  Processing Systems\/}, NIPS'17 (Curran Associates Inc., USA, 2017).

\bibitem{umls}
O.~Bodenreider, The unified medical language system (umls): integrating
  biomedical terminology, {\em Nucleic Acids Research} {\bf 32}, D267  (2004).

\bibitem{gensim}
R.~{\v R}eh{\r u}{\v r}ek and P.~Sojka, {Software Framework for Topic Modelling
  with Large Corpora}, in {\em {Proceedings of the LREC 2010 Workshop on New
  Challenges for NLP Frameworks}\/},  (ELRA, Valletta, Malta, May 2010).

\bibitem{BinaryRelevance}
G.~Varando, C.~Bielza and P.~Larra{\~{n}}aga, Expressive power of binary
  relevance and chain classifiers based on bayesian networks for multi-label
  classification, in {\em Probabilistic Graphical Models\/},  eds. L.~C.
  van~der Gaag and A.~J. Feelders (Springer International Publishing, Cham,
  2014).

\bibitem{GELU}
D.~Hendrycks and K.~Gimpel, Bridging nonlinearities and stochastic regularizers
  with gaussian error linear units, {\em CoRR} {\bf abs/1606.08415}  (2016).

\bibitem{hierarchy}
F.~Morin and Y.~Bengio, Hierarchical probabilistic neural network language
  model, in {\em AISTATS\/}, 2005.

\bibitem{Radford2018ImprovingLU}
A.~Radford, Improving language understanding by generative pre-training2018.

\bibitem{miotto2016deep}
R.~Miotto, L.~Li, B.~A. Kidd and J.~T. Dudley, Deep patient: an unsupervised
  representation to predict the future of patients from the electronic health
  records, {\em Scientific reports} {\bf 6}, p. 26094  (2016).

\bibitem{ELECTRA}
K.~Clark, M.-T. Luong, Q.~V. Le and C.~D. Manning, Electra: Pre-training text
  encoders as discriminators rather than generators, in {\em International
  Conference on Learning Representations\/}, 2020.

\end{thebibliography}

\clearpage

\appendix

\section{Language Model Hyperparameter Grid} \label{lm_grid}

\begin{table}[H]
\begin{center}
\caption{Language Model Hyperparameters}
{\begin{tabular}{cc}\toprule
Hyperparameter Name & Hyperparameter Values \\ \hline
Embedding Size & [400, 800] \\
GRU Hidden Size & [400, 800, 1600] \\
LR & $[10^{-2}, 10^{-3}, 10^{-4}, 10^{-5}]$ \\
$L_2$ & [0.1, 0.01, 0.001] \\
Dropout & [0, 0.1, 0.2] \\
\hline
\end{tabular}}
\end{center}
\end{table}

\section{Best Language Model Hyperparameters} \label{Best_lm}

\begin{table}[H]
\begin{center}
\caption{Best Language Model Hyperparameters}
{\begin{tabular}{ccc}\toprule
Hyperparameter Name & Size 400 Model Value & Size 800 Model Value \\ \hline
Embedding Size & 400 & 800 \\
GRU Hidden Size & 800 & 1600 \\
LR & $10^{-3} $ & $10^{-3}$  \\
$L_2$ & 0.01 & 0.1 \\
Dropout & 0.1 & 0.1 \\
Epochs & 20 & 40 \\
\hline
\end{tabular}}
\end{center}
\end{table}

\section{End-to-end GRU Model Hyperparameter Grid} \label{end_to_end_grid}

\begin{table}[H]
\begin{center}
\caption{End-to-end GRU Model Model Hyperparameters}
{\begin{tabular}{cc}\toprule
Hyperparameter Name & Hyperparameter Values \\ \hline
Embedding Size & [100, 200, 400] \\
GRU Hidden Size & [100, 200, 400] \\
LR & $[10^{-2}, 10^{-3}, 10^{-4}, 10^{-5}]$ \\
$L_2$ & [0.1, 0.01, 0.001] \\
Dropout & [0, 0.1, 0.2] \\
\hline
\end{tabular}}
\end{center}
\end{table}

\section{Best End-to-end GRU Model Hyperparameters} \label{Best_gru}

\begin{table}[H]
\begin{center}
\caption{Inpatient Mortality GRU Best Hyperparameters}
{\begin{tabular}{cccc}\toprule
Hyperparameter Name &  Value \\ \hline
Embedding Size & 100 \\
GRU Hidden Size & 400 \\
LR & $10^{-2} $ \\
$L_2$ & 0.1\\
Dropout & 0.1 \\
Epochs & 21 \\
\hline
\end{tabular}}
\end{center}
\end{table}

\begin{table}[H]
\begin{center}
\caption{Long Admission GRU Best Hyperparameters}
{\begin{tabular}{cccc}\toprule
Hyperparameter Name &  Value \\ \hline
Embedding Size & 400 \\
GRU Hidden Size & 100 \\
LR & $10^{-2} $ \\
$L_2$ & 0.1\\
Dropout & 0.1 \\
Epochs & 28 \\
\hline
\end{tabular}}
\end{center}
\end{table}

\begin{table}[H]
\begin{center}
\caption{ICU Transfer GRU Best Hyperparameters}
{\begin{tabular}{cccc}\toprule
Hyperparameter Name &  Value \\ \hline
Embedding Size & 400 \\
GRU Hidden Size & 400 \\
LR & $10^{-3} $ \\
$L_2$ & 0.001 \\
Dropout & 0 \\
Epochs & 0 \\
\hline
\end{tabular}}
\end{center}
\end{table}

\begin{table}[H]
\begin{center}
\caption{30-day Readmission GRU Best Hyperparameters}
{\begin{tabular}{cccc}\toprule
Hyperparameter Name &  Value \\ \hline
Embedding Size & 400 \\
GRU Hidden Size & 100 \\
LR & $10^{-2} $ \\
$L_2$ & 0.1 \\
Dropout & 0 \\
Epochs & 24 \\
\hline
\end{tabular}}
\end{center}
\end{table}

\begin{table}[H]
\begin{center}
\caption{Abnormal HbA1c GRU Best Hyperparameters}
{\begin{tabular}{cccc}\toprule
Hyperparameter Name &  Value \\ \hline
Embedding Size & 400 \\
GRU Hidden Size & 200 \\
LR & $10^{-3} $ \\
$L_2$ & 0.01 \\
Dropout & 0.1 \\
Epochs & 1 \\
\hline
\end{tabular}}
\end{center}
\end{table}

\section{Best Prediction Model/Representation Hyperparameters On All Data} \label{Best_task}

\begin{table}[H]
\caption{Inpatient Mortality Best Hyperparameters}
{\begin{tabular}{cccp{5cm}}\toprule
Representation Name & Representation Hyperparameters & Best Model Type & Best Hyperparameters \\ \hline
Counts & with\_ontology\_expansion & LightGBM & num\_leaves: 100 \newline num\_boost\_round: 317 \newline learning\_rate: 0.02 \\
Word2Vec & concat\_max\_mean\_min & Logistic & C: 0.01 \\
LSI & size: 800 & LightGBM & num\_leaves: 10 \newline num\_boost\_round: 250 \newline learning\_rate: 0.02 \\
CLMBR & size: 800 & Logistic & C: 0.001 \\
\hline
\end{tabular}}
\end{table}

\begin{table}[H]
\caption{Long Admission Best Hyperparameters}
{\begin{tabular}{cccp{5cm}}\toprule
Representation Name & Representation Hyperparameters & Best Model Type & Best Hyperparameters \\ \hline
Counts & with\_time\_bins & LightGBM & num\_leaves: 100 \newline num\_boost\_round: 292 \newline learning\_rate: 0.02 \\
Word2Vec & concat\_max\_mean\_min & LightGBM & num\_leaves: 100 \newline num\_boost\_round: 360 \newline learning\_rate: 0.02 \\
LSI & size: 800 & LightGBM & num\_leaves: 100 \newline num\_boost\_round: 494 \newline learning\_rate: 0.02 \\
CLMBR & size: 800 & LightGBM & num\_leaves: 100 \newline num\_boost\_round: 397 \newline learning\_rate: 0.02 \\
\hline
\end{tabular}}
\end{table}

\begin{table}[H]
\caption{ICU Transfer Best Hyperparameters}
{\begin{tabular}{cccp{5cm}}\toprule
Representation Name & Representation Hyperparameters & Best Model Type & Best Hyperparameters \\ \hline
Counts & with\_time\_bins & LightGBM & num\_leaves: 100 \newline num\_boost\_round: 43 \newline learning\_rate: 0.02 \\
Word2Vec & with\_ontology\_expansion,concat\_max\_mean\_min & Logistic & C: 1.0 \\
LSI & size: 800 & Logistic & C: 1000000.0 \\
CLMBR & size: 800 & Logistic & C: 1e-05 \\
\hline
\end{tabular}}
\end{table}

\begin{table}[H]
\caption{30-day Readmission Best Hyperparameters}
{\begin{tabular}{cccp{5cm}}\toprule
Representation Name & Representation Hyperparameters & Best Model Type & Best Hyperparameters \\ \hline
Counts & with\_time\_bins & LightGBM & num\_leaves: 100 \newline num\_boost\_round: 159 \newline learning\_rate: 0.02 \\
Word2Vec & concat\_max\_mean\_min & LightGBM & num\_leaves: 100 \newline num\_boost\_round: 215 \newline learning\_rate: 0.02 \\
LSI & size: 400 & LightGBM & num\_leaves: 100 \newline num\_boost\_round: 188 \newline learning\_rate: 0.02 \\
CLBMR & size: 800 & LightGBM & num\_leaves: 100 \newline num\_boost\_round: 282 \newline learning\_rate: 0.02 \\
\hline
\end{tabular}}
\end{table}

\begin{table}[H]
\caption{Abnormal HbA1c Best Hyperparameters}
{\begin{tabular}{cccp{5cm}}\toprule
Representation Name & Representation Hyperparameters & Best Model Type & Best Hyperparameters \\ \hline
Counts & with\_ontology\_expansion & LightGBM & num\_leaves: 100 \newline num\_boost\_round: 73 \newline learning\_rate: 0.1 \\
Word2Vec & concat\_max\_mean\_min & LightGBM & num\_leaves: 25 \newline num\_boost\_round: 21 \newline learning\_rate: 0.1 \\
LSI & size: 800 & LightGBM & num\_leaves: 10 \newline num\_boost\_round: 63 \newline learning\_rate: 0.1 \\
CLMBR & size: 800 & Logistic & C: 0.01 \\
\hline
\end{tabular}}
\end{table}

\end{document}